\documentclass[10pt,twocolumn,letterpaper]{article}

\usepackage[pagenumbers]{cvpr} 

\definecolor{cvprblue}{rgb}{0.21,0.49,0.74}
\usepackage[pagebackref,breaklinks,colorlinks,allcolors=cvprblue]{hyperref}
\usepackage{colortbl}
\usepackage{pifont}       
\usepackage{bbding}       
\usepackage{fontawesome}  

\def\paperID{8317} 
\def\confName{CVPR}
\def\confYear{2025}

\definecolor{Gray}{gray}{0.95}
\definecolor{barriercolor}{RGB}{255, 120, 50}
\definecolor{bicyclecolor}{RGB}{255, 192, 203}
\definecolor{buscolor}{RGB}{255, 255, 0}
\definecolor{carcolor}{RGB}{0, 150, 245}
\definecolor{constructcolor}{RGB}{0, 255, 255}
\definecolor{motorcolor}{RGB}{200, 180, 0}
\definecolor{pedestriancolor}{RGB}{255, 0, 0}
\definecolor{trafficcolor}{RGB}{255, 240, 150}
\definecolor{trailercolor}{RGB}{135, 60, 0}
\definecolor{truckcolor}{RGB}{160, 32, 240}
\definecolor{driveablecolor}{RGB}{255, 0, 255}
\definecolor{otherflatcolor}{RGB}{139, 137, 137}
\definecolor{sidewalkcolor}{RGB}{75, 0, 75}
\definecolor{terraincolor}{RGB}{150, 240, 80}
\definecolor{manmadecolor}{RGB}{213, 213, 213}
\definecolor{vegetationcolor}{RGB}{0, 175, 0}
\definecolor{otherscolor}{RGB}{0, 0, 0}

\title{STCOcc: Sparse Spatial-Temporal Cascade Renovation for 3D Occupancy and Scene Flow Prediction}

\author{Zhimin Liao, ~Ping Wei\textsuperscript{\rm}\thanks{Corresponding author.}, ~Shuaijia Chen, ~Haoxuan Wang, ~Ziyang Ren\\
National Key Laboratory of Human-Machine Hybrid Augmented Intelligence\\
Institute of Artificial Intelligence and Robotics, Xi’an Jiaotong University\\
{\tt\small liaozm@stu.xjtu.edu.cn,~pingwei@xjtu.edu.cn}}

\begin{document}
\maketitle
\begin{abstract}
    3D occupancy and scene flow offer a detailed and dynamic representation of 3D scene. Recognizing the sparsity and complexity of 3D space, previous vision-centric methods have employed implicit learning-based approaches to model spatial and temporal information. However, these approaches struggle to capture local details and diminish the model's spatial discriminative ability. To address these challenges, we propose a novel explicit state-based modeling method designed to leverage the occupied state to renovate the 3D features. Specifically, we propose a sparse occlusion-aware attention mechanism, integrated with a cascade refinement strategy, which accurately renovates 3D features with the guidance of occupied state information. Additionally, we introduce a novel method for modeling long-term dynamic interactions, which reduces computational costs and preserves spatial information. Compared to the previous state-of-the-art methods, our efficient explicit renovation strategy not only delivers superior performance in terms of RayIoU and mAVE for occupancy and scene flow prediction but also markedly reduces GPU memory usage during training, bringing it down to 8.7GB. Our code is available on \url{https://github.com/lzzzzzm/STCOcc}
    
\end{abstract}    
\begin{figure}[t!]
    \centering
    \begin{subfigure}{0.47\textwidth}
        \centering
        \includegraphics[width=\linewidth]{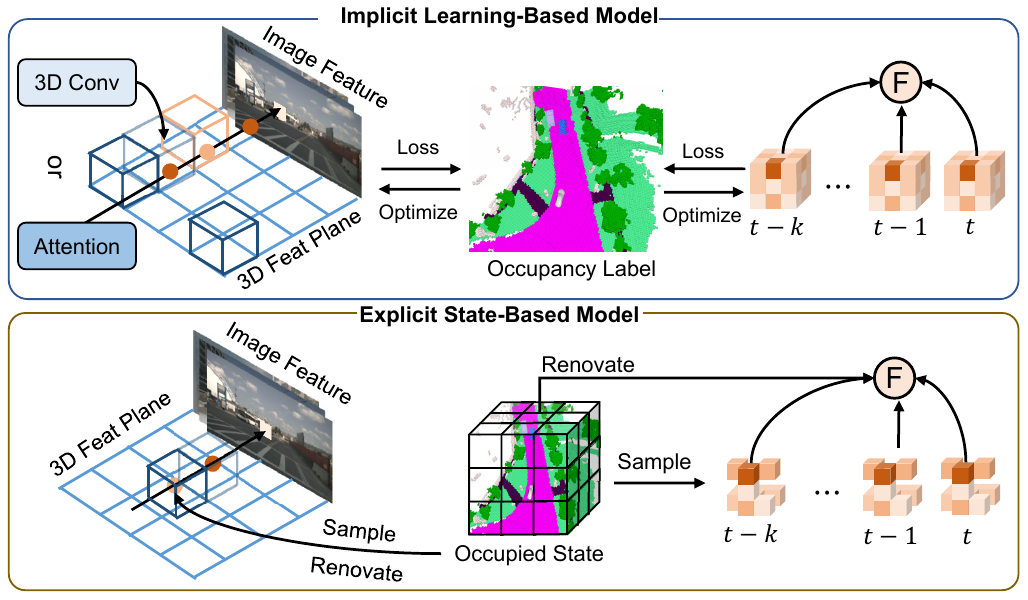}
        \caption{Explicit versus Implicit Modeling.}
        \label{figure:motivation-Spatial-modeling}
    \end{subfigure}
    \centering
    \begin{subfigure}{0.47\textwidth}
        \centering
        \includegraphics[width=\linewidth]{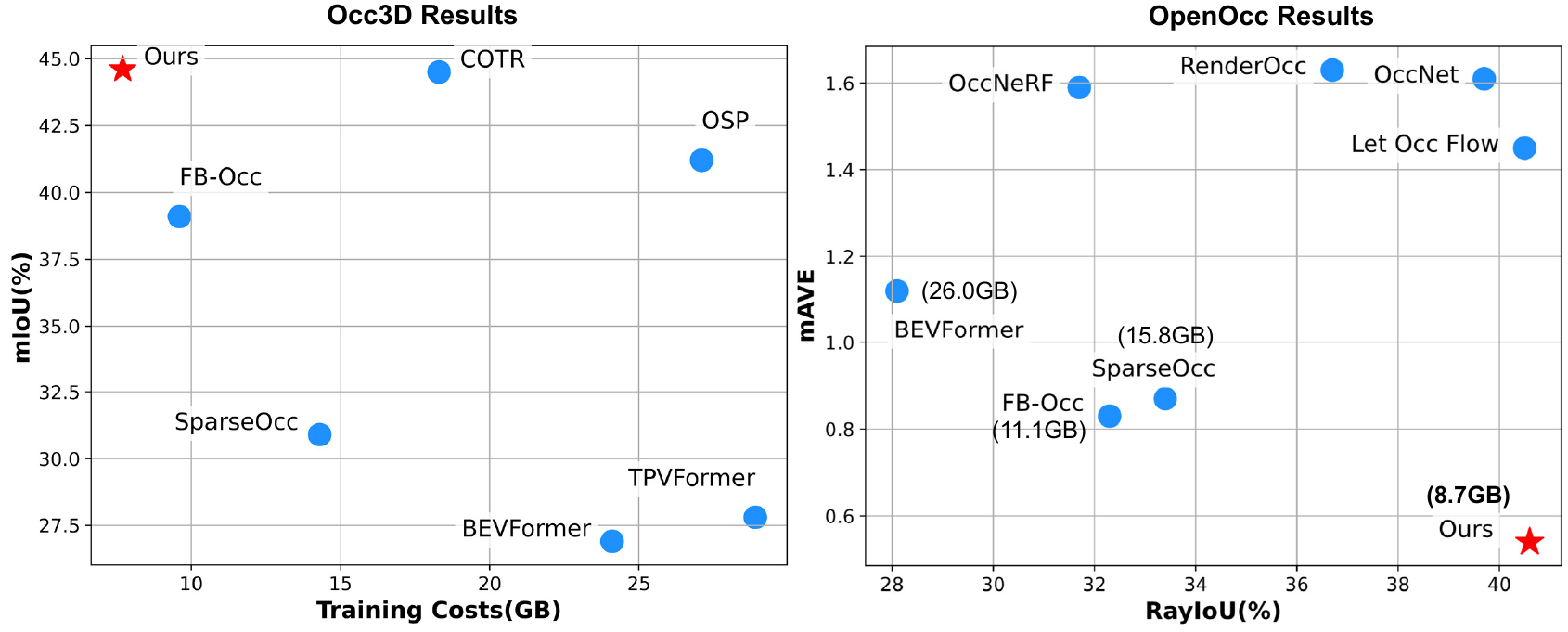}
        \caption{Comparison with Different Methods.}
        \label{figure:motivation-model-comparsion}
    \end{subfigure}
    \caption{\textbf{(a) Explicit versus Implicit Modeling:} We propose a novel explicit state-based modeling approach that explicitly leverages the occupied state to maintain feature sparsity and model spatial details. \textbf{(b) Comparison with Different Methods:} Our approach achieves state-of-the-art performance of RayIoU and mAVE with lower training costs.}
    \vspace{-1.5em}
    \label{figure:motivation}
\end{figure}

\section{Introduction}

    Accurate perception of 3D surrounding scenes is indeed vital for autonomous systems. The goal of occupancy and scene flow prediction~\cite{tesla_ai_day} is to segment the entire space into 3D voxels and to determine the semantic and flow information of each voxel. This capability is crucial for understanding the environment and making informed decisions, which is well-suited for downstream tasks in autonomous systems, such as mapping and planning~\cite{occ3d, occnet, semantickitti, nuscenes, vidar}.
    
    Due to data sparsity and information redundancy in 3D space, employing efficient and robust approaches for 3D feature processing is critical. Existing vision-centric methods~\cite{bevformer, bevdet, occformer, tpvformer, surroundocc, occ3d, occdepth, fb-occ, sparseocc, voxformer, sgn} rely on implicit learning-based method for modeling spatial and temporal information, as shown in \cref{figure:motivation-Spatial-modeling}. While these methods optimize 3D reconstruction and long-term temporal fusion through loss supervision, they face inherent limitations. Specifically, the exclusive reliance on loss-driven training hinders the model’s ability to capture fine-grained spatial details and compromises the spatial discriminative capacity of learned features, ultimately reducing the effectiveness for occupancy prediction tasks.  

     In this paper, we propose an explicit state-based modeling method that leverages the occupied state of 3D space to refine spatial and temporal feature representations, as shown in \cref{figure:motivation-Spatial-modeling}. Our key insight stems from the inherent geometric correspondence between the occupied state and the 3D structure. Since the occupied state directly encodes the geometry of 3D space, it can serve as a robust prior to guide feature learning. This geometric alignment ensures that the feature space preserves structural fidelity, thereby simplifying the learning process and enhancing the discriminative power of the model. 

     We introduce STCOcc, a sparse \textbf{S}patial-\textbf{T}emporal \textbf{C}ascade renovation framework tailored for occupancy and scene flow prediction. Within this framework, we present a \textit{Spatial-Temporal Cascade Decoder} designed to renovate the 3D features both spatially and temporally, leveraging the occupied state of the 3D space. Specifically, at each stage, we employ a \textit{Self-Recursive Occupancy Predictor} (SROP) to progressively refine the occupied state of the 3D space, thereby providing a more precise 3D geometric state for renovating the 3D features. Subsequently, we propose a sparse occlusion-aware attention mechanism to renovate the 3D features. Our attention mechanism differs from prior methods \cite{sparse4d, fbbev, bevformer}, which relied solely on depth information to renovate the 3D features. Instead, we utilize the occupied state in conjunction with bin depth information to accurately model the 3D spatial features. This approach provides details of local regions and makes the features more geometrically accurate. Furthermore, leveraging the accurate occupied state identified by SROP, we employ the \textit{Occlusion-Aware Temporal Self-Attention} (OA-TSA) to model dynamic information using a recurrent strategy, supplying detailed short-term temporal information.

     To efficiently integrate long-term temporal fusion, we propose a novel sparse-based method for temporal fusion modeling. It also avoids redundant information in historical data and preserves spatial information. Specifically, based on the occupied state, we sample non-empty and empty regions into long-term and short-term streams, respectively. Then, we incorporate the occupied state into both streams and employ a parallel strategy to fuse the temporal information within these two streams. This approach not only reduces computational costs but also retains the spatial information within the 3D space. Our contributions can be summarized as follows: 

    \begin{itemize}

        \item We introduce an explicit state-based modeling approach designed to renovate the 3D features both spatially and temporally.

        \item We propose a sparse, occlusion-aware mechanism that provides more accurate geometric 3D features. Additionally, we propose a novel sparse-based method for modeling long-term dynamic information. This approach not only reduces computational costs but also ensures spatial consistency.

        \item Our method achieves a RayIoU of 41.7$\%$ on Occ3D~\cite{occ3d} and a RayIoU of 40.8$\%$ along with a mAVE of 0.44 for occupancy and scene flow prediction on OpenOcc~\cite{occnet}, while also reducing the training memory usage to 8.7GB, as shown in \cref{figure:motivation-model-comparsion}.
        
    \end{itemize}
\section{Related Work}
    \subsection{Camera-based 3D Occupancy Prediction}
        Occupancy, as proposed by~\cite{occupancy-network, conv-occupancy}, focuses on the continuous representation of 3D scenes. MonoScene \cite{monoscene} leverages monocular images for semantic scene completion, employing a 3D UNet to process voxel features. TPVFormer \cite{tpvformer} lifts image features into 3D TPV space and expands them into voxel representations for 3D occupancy prediction. OccFormer \cite{occformer} proposes a dual-path transformer for encoding the dense 3D volume features. 

        Considering the inherent sparsity of 3D scenes, recent methods~\cite{voxformer, vampire, sparseocc, sgn, sparseocc-cvpr, cascadeflow} optimize computational efficiency by processing only non-empty voxels using sparse convolutions or attention mechanisms. VoxFormer \cite{voxformer} utilizes a depth-based query proposal network to generate sparse query proposals for 3D-to-2D cross-attention. SGN \cite{sgn} introduces a dense-sparse-dense framework that dynamically selects sparse seed voxels and employs hybrid guidance to enhance the convergence of semantic diffusion. Symphonize \cite{symphonize} reconstructs the 3D scene using instance queries. SparseOcc \cite{sparseocc} proposes a fully sparse framework that focuses exclusively on non-empty regions. However, these sparse methods rely solely on the occupied state of 3D space to select the region of interest which ignores feature semantics or contextual relationships. Several methods~\cite{occupancym3d, vampire} leverage occupancy-based loss supervision to refine 3D features, improving the spatial fidelity of feature representations. However, their exclusive reliance on loss-driven optimization restricts their ability to model fine-grained 3D spatial structures.

    \begin{figure*}[t]
    \centering
    \includegraphics[width=0.90\textwidth]{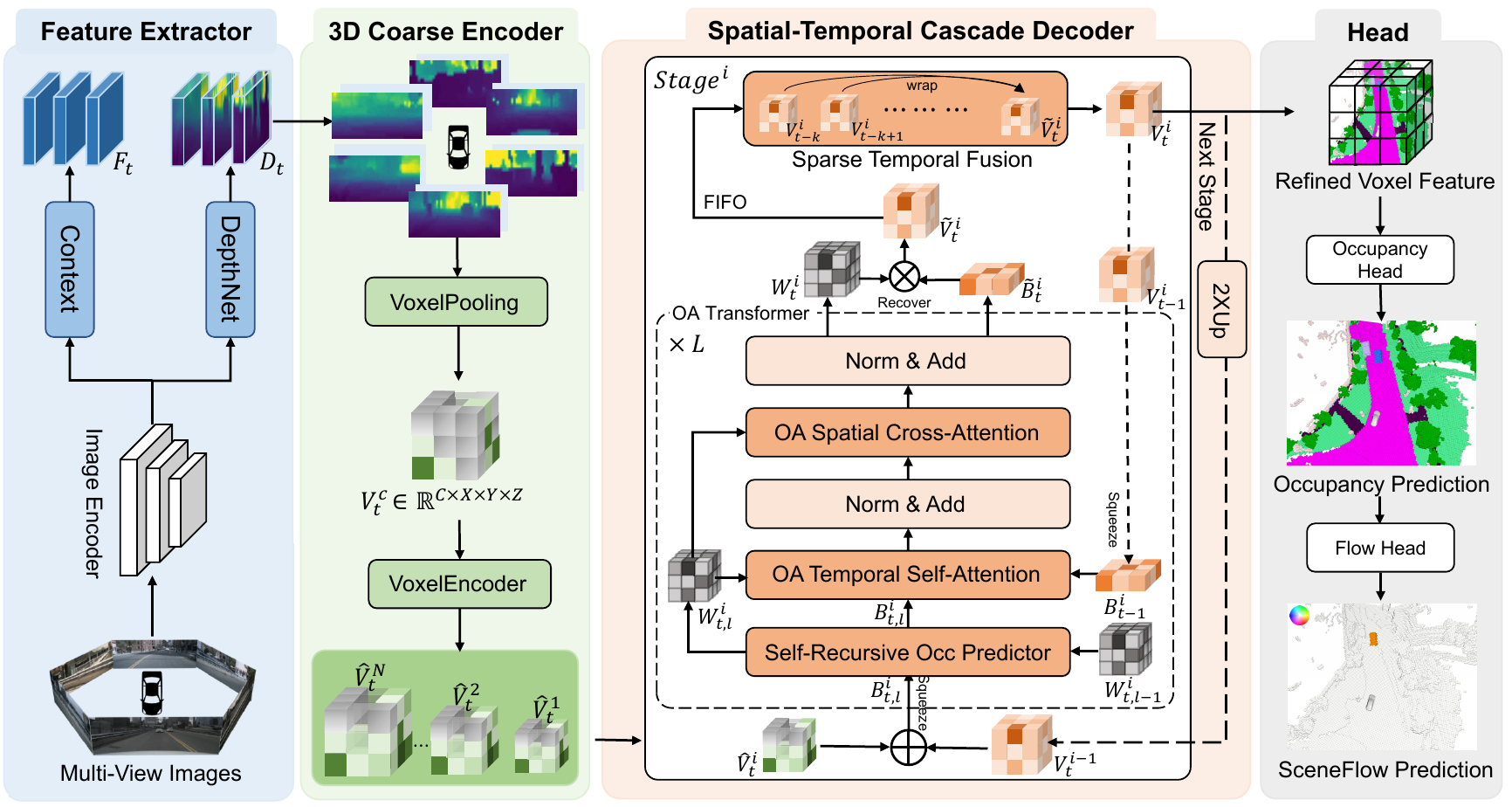}
    \vspace*{-0.1cm}
    \caption{ 
    \textbf{The overall architecture of STCOcc.} The STCOcc framework is primarily composed of four integral modules: a feature extractor that captures image features and depth distribution, a 3D coarse encoder that generates multi-resolution coarse voxel features, a multi-stage spatial-temporal cascade decoder that incrementally renovates these coarse voxel features in both spatial and temporal dimensions, and a head module designed to leverage the refined voxel features for the prediction of 3D occupancy and scene flow.
    }
    \label{label:model_framework}
    \vspace{-1em}
\end{figure*}

    \subsection{Camera-based Temporal Modeling}
        Temporal modeling is essential for camera-based perception due to the inherent challenge of lacking depth information. When considering the modeling space, the methods can be divided into two main categories: image feature-based~\cite{petr, petrv2, sparse4d, sparsebev, detr3d} and 3D feature-based~\cite{bevformer, bevformerv2, occnet, solofusion, bevstereo, bevdepth, bev-lanedet, bevverse}. Image feature-based temporal modeling methods utilize multi-frame image features to provide dynamic information. For instance, PETR~\cite{petr, petrv2} projects 3D points onto multi-frame image features to generate implicit 3D features for modeling temporal information. Sparse4D~\cite{sparse4d} creates 4D keypoints based on 3D anchors and projects these points to aggregate features from multi-frame image features. SparseBEV~\cite{sparsebev} adaptively generates sampling points based on the query features.
        
        On the other hand, 3D feature-based methods model temporal features in BEV or voxel space. BEVFormer~\cite{bevformer} designs a temporal self-attention mechanism to recursively fuse BEV features. OccNet~\cite{occnet} extends this paradism to voxel-based temporal self-attention to recursively fuse voxel feature. SOLOFusion~\cite{solofusion} uses a parallel strategy to model BEV-level long-term information. We propose a novel sparse approach to model long-term 3D features.
\section{Methods}

    \subsection{Overall Architecture}
        \label{label:Overall Architecture}

        An overview of STCOcc is presented in \cref{label:model_framework}. Along the timestamp, we take the multi-view images as video sequence. At current frame $t$, multi-view images are first processed by the \textit{Feature Extractor} to obtain image features and depth distribution. The \textit{3D Coarse Encoder} uses the image features and depth distribution to create multi-resolution coarse voxel features via LSS-based transformation \cite{lss, bevdepth, bevstereo}. The \textit{Spatial-Temporal Cascade Decoder} then progressively renovates these voxel features spatially and temporally, stage by stage. Finally, the \textit{Head} module utilizes the refined voxel feature to predict occupancy and scene flow.
        
        \vspace{-1em}
        \paragraph{Feature Extractor.} At each time $t$, the feature extractor initially uses an image backbone (e.g., ResNet \cite{resnet}) to extract multi-view features \( F_t = \{ F_t^{j} \in \mathbb{R}^{C \times H \times W} \}_{j=1}^{N_{c}} \), where \( F_t^{j} \) represents the features of the \( j \)-th camera view at time $t$, and \( N_{c} \) is the number of cameras. Then, the depth network~\cite{bevdepth, bevstereo} processes these image features to predict the bin depth distribution \( D_t = \{ D_t^{j} \in \mathbb{R}^{D_{bin} \times H \times W} \}_{j=1}^{N_{c}} \).
        \vspace{-1em}
        \paragraph{3D Coarse Encoder.} The 3D coarse encoder adheres to the Lift and Splat framework as delineated in the LSS paradigm \cite{lss, bevdepth, bevstereo}. In the Lift phase, each pixel within the 2D image feature planes \( F_t \) is projected into the 3D voxel space guided by the predicted bin depth distribution \( D_t \). Subsequently, the Splat phase consolidates the feature values of pixels falling within each voxel through voxel pooling \cite{lss, bevdepth, bevstereo}, thereby constructing the coarse voxel feature \( V_t^{c} \in \mathbb{R}^{C \times X \times Y \times Z} \). Subsequently, we engage a lightweight voxel encoder (e.g. ResNet3D-18 \cite{resnet}) to produce multi-resolution coarse voxel features \( V_t^{c} = \{ \hat{V}_t^i \in \mathbb{R}^{C \times X_{i} \times Y_{i} \times Z_{i}} \mid i = 1, 2, \ldots, N \} \), where \( X_{i} = \frac{X}{2^{(N-i)}}, Y_{i} = \frac{Y}{2^{(N-i)}}, Z_{i} = \frac{Z}{2^{(N-i)}} \), and \( N \) is the number of processing stages.

    \subsection{Spatial-Temporal Cascade Decoder}
        \label{label:Spatial-Temporal Cascade Decoder}

        To explicitly model the spatial and temporal information into features with the occupied state of 3D space, we introduce a spatial-temporal cascade decoder that renovates the 3D coarse voxel features $V_t^{c}$ through a multi-stage process. As depicted in \cref{label:model_framework}, the decoder comprises two primary components: Occlusion-Aware (OA) transformer layers, which accurately capture spatial and short-term temporal information, and a sparse-based temporal fusion module that employs a first-in, first-out (FIFO) memory sequence to encode long-term temporal information.

        At each stage $i$, we refine voxel feature in BEV space rather than in voxel space. Initially, we fuse the coarse voxel feature $\hat{V}_t^{i}$ with the refined voxel feature from the previous stage output $V_{t}^{i-1}$, and then project it into the BEV representation to obtain $B^{i}_{t,0} \in \mathbb{R}^{C\times X_{i}\times Y_{i}}$ as the input of OA transformer. We subsequently apply $L$ layers \textit{OA-Transformer}, which is analogous to the approach in \cite{bevformer}, to refine $B^{i}_{t,0}$. This transformer layer includes three specialized modules: the \textit{Self-Recursive Occupancy Predictor} (SROP), \textit{OA Temporal Self-Attention} (OA-TSA), and \textit{OA Spatial Cross-Attention} (OA-SCA). The SROP is designed to provide an accurate occupied state of the 3D space for each stage. The OA-TSA captures short-range temporal dynamics within the BEV space. The OA-SCA explicitly utilizes the occupied state to address the ambiguous projection problem~\cite{bevformer, fbbev}, which transfers geometric 2D feature information into the 3D space.
        
        After the \textit{OA-Transformer} processes the features, the refined BEV features $\tilde{B}_{t}^{i}$ are converted back to voxel form $\tilde{V}_{t}^{i}$ using the occupied weight $W_{t}^{i}$. Subsequently, we leverage the occupied state of the 3D space to guide a novel sparse-based approach for modeling long-term temporal information, thereby obtaining the output $V_{t}^{i}$ at time $t$. $V_{t}^{i}$ is upsampled by a factor of 2 for the next stage of refinement.

            \subsubsection{Self-Recursive Occupancy Predictor}
                \label{label:Self-Recursive Occupancy Predictor}
                To provide a more accurate representation of the occupied state of 3D space and to mitigate the one-off selection issues present in previous methods \cite{voxformer, sparseocc, sgn}. We draw inspiration from earlier studies \cite{heightformer, sparse4d, widthformer} and design the self-recursive occupancy predictor. This predictor employs successive transformer layers to progressively refine the occupied state layer by layer. Specifically, at each layer $l$, it utilizes a simple Multilayer Perceptron (MLP) to recover the height of \( B_{t,l}^{i} \) to voxel space and predict the occupancy weights \( W_{t,l}^{i} \in \mathbb{R}^{X_{i} \times Y_{i}\times Z_{i}} \). The process of the self-recursive occupancy predictor can be described as follows:
                \begin{equation}
                    W_{t,l}^{i} = f^{i}(B_{t,l}^{i}) + \alpha_{l}^{i} W_{t,l-1}^{i},
                \end{equation}
                The function \( f^{i}(\cdot) \) corresponds to the occupancy predictor in stage \( i \), which shares the same weights across different layers. The parameter \( \alpha_{l}^{i} \) signifies the effect of layer \( l-1 \), a learnable parameter initialized to 0.5. At each stage, the initial \( W_{t,0}^{i} \) is derived by upsampling the occupied weights from the previous stage, except that it is set to zeros in the first stage.

            \subsubsection{OA Temporal Self-Attention}
                \label{label:OA Temporal Self-Attention}
                The temporal modeling is crucial for representing the dynamic driving scene \cite{bevformer}. Given the historical BEV feature \( B_{t-1}^{i} \)(discussed in \cref{label:Long-Term Modeling}), we align it with the current feature \( B_{t,l}^{i} \) via the motion of the ego-vehicle. To efficiently model the dynamic information, we propose Occlusion-Aware Temporal Self-Attention (OA-TSA) to focus the temporal modeling on the non-empty space. The OA-TSA is represented by:
                \begin{small}
                \begin{equation}
                    TSA_{OA}(Q_{x,y}, \mathcal{B}, \overline{W}) = \sum_{b \in \mathcal{B}} \overline{w}_{x,y} \mathcal{F}_{d}(Q_{x,y}, b, \overline{w}_{x,y}),
                \end{equation}
                \end{small}
                where \( Q_{x,y} \) denotes the BEV feature located at \( p = (x, y) \) and \( \mathcal{B} = \{B_{t,l}^{i}, B_{t-1}^{i}\} \).\( \overline{W} \in \mathbb{R}^{X_{i} \times Y_{i}} \) is the average of \( W_{t,l}^{i} \) along the z-axis, and \( \mathcal{F}_{d} \) signifies the deformable attention mechanism \cite{deformable-detr}. \( \overline{w}_{x,y} \) represents the occupied weights $\overline{W}$ at position \( p \). Unlike the vanilla deformable attention~\cite{deformable-detr}, the offsets are predicted by the concatenation of \( \overline{W} \) and \( \mathcal{B} \). By reweighting the TSA, we enable the model to focus more effectively on the dynamic information within the 3D space.
            
            \subsubsection{OA Spatial Cross-Attention}
                \label{label:OA Spatial Cross-Attention}
                To explicitly utilize the occupied state to renovate the 3D features, we propose the Occupancy-Aware Spatial Cross-Attention (OA-SCA), which leverages the occupied state to enhance geometric features. 
                
\begin{figure}[tb]
    \centering
    \includegraphics[width=0.47\textwidth]{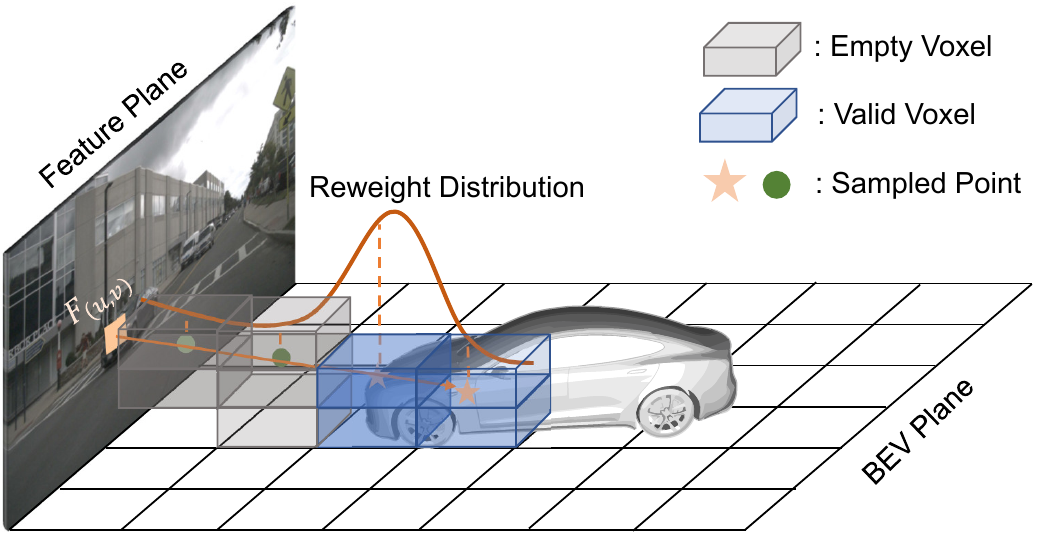}
    \vspace*{-0.1cm}
    \caption{ 
    \textbf{Illustration of OA-SCA.} Due to the projection process, sampled points along the same ray in the feature plane are assigned identical features, even when they represent empty voxel space, as depicted by the green points. To address this limitation, our approach integrates depth and occupancy information to assign appropriate weights to the sampled points, thereby enhancing the differentiation of features along each ray.
    }
    \label{label:Spatia-Cross-Attention}
    \vspace{-1.5em}
\end{figure}

                We first revisit the vanilla Spatial Cross-Attention (SCA)~\cite{bevformer} as follows. As shown in \cref{label:Spatia-Cross-Attention}, it samples $N_{ref}$ 3D points $\textbf{X}=\{\textbf{x}=(x,y,z_{h})|h=1,2 \cdots, N_{ref}\}$ with different height for each $Q_{x,y}$, and projects these 3D points to 2D image feature planes $F_{t}$ to obtain corresponding features. Formally, the SCA can be expressed as:
                \vspace{-0.5em}
                \begin{equation}
                \begin{split}
                    SCA(Q_{x,y}, F_{t}) &= \sum_{\textbf{x} \in \textbf{X}\textbf{}} \mathcal{F}_{d}(Q_{x,y}, P(\textbf{x}), F_{t}),
                \end{split}
                \end{equation}
                To simplify our initial analysis, we consider a scenario with a single camera, \( P(\cdot) \) is the projection matrix that projects points from the 3D space onto the feature plane. The projection process can be mathematically represented as:
                \begin{equation}
                \label{label:equation-projection}
                    d \cdot
                    \left[
                    \begin{array}{ccc}
                         u & v & 1 \\
                    \end{array}
                    \right]^{T} = P \cdot \left[
                    \begin{array}{cccc}
                         x & y & z & 1 \\
                    \end{array}
                    \right]^{T},
                \end{equation}
                where \( d \) denotes the depth of the point \((u, v)\) on the 2D image plane. This 3D to 2D transformation introduces ambiguity, as different 3D points along the same projection ray map to identical 2D coordinates and are assigned the same features $F_{(u,v)}$, as illustrated in \cref{label:Spatia-Cross-Attention}, where even the green points corresponding to empty areas receive the same feature.

                To address this ambiguity, previous methods \cite{fbbev, sparse4d} propose utilizing depth information to reweight the features of sampled points. However, these methods overlook the precision of the predicted depth and the state attributes of the sampled features. In contrast, our approach is inspired by volume rendering techniques \cite{volumerendering, nerf}, which allows us to renovate the features of sampled points more effectively. The volume rendering can be represented as:
                \begin{equation}
                    \mathcal{C} = \sum_{n=1}^{N_{r}} \tau_{n} \cdot \sigma_{n} \cdot c_{n},
                \end{equation}
                where \(\mathcal{C} \) represents the expected value of light emitted by particles within the volume as a ray samples \( N_{r} \) points. \( \sigma_{n} \) denotes the density at each point, \( \tau_{n} \) is the transmittance, and \( c_{n} \) is the corresponding color value. This physical function is analogous to the reverse processing of SCA, where \( \mathcal{C} \) corresponds to \( F_{(u,v)} \) in the image plane, and \( c_{n} \) corrresponds to the sampled point features, as illustrated in \cref{label:Spatia-Cross-Attention}.
                
                Based on above observation, our propose OA-SCA is designed to address the ambiguity inherent in vanilla SCA. Furthermore, to maintain the sparsity of the model, we employ probability sampling to select 3D reference points for refinement. The OA-SCA can be formulated as:
                \begin{equation}
                    SCA_{OA}(Q_{x,y}, F_{t}) = \sum_{\textbf{x} \in \textbf{X}_{s}} \Omega_{\textbf{x}} \mathcal{F}_{d}(Q_{x,y}, P(\textbf{x}), F_{t}),
                \end{equation}
                where \( \textbf{X}_{s} = \{\textbf{x} \in \textbf{X} \mid w_{\textbf{x}} > u_{\textbf{x}} \} \). During training, \( u_{\textbf{x}} \) follows a truncated normal distribution \((0.5, 1)\). For stable inference, \( u_{\textbf{x}} \) is set to 0.5. This sampling approach allows our method to account for the uncertainty region during training. \( w_{\textbf{x}} \) is the reference point \( \textbf{x} \) corresponding occupied weight. The reweighting parameter \( \Omega_{\textbf{x}} = w_{\textbf{x}} \cdot \beta_{\textbf{x}} \), for each 3D point $\textbf{x}$, \( \beta_{\textbf{x}} \) can be calculated as:
                \begin{small}
                \begin{equation}
                    \beta_{\textbf{x}} = \exp\left(-\frac{\min\left(\left|d_{r} - (d_{r}^{\prime} - \Delta d)\right|, \left|d_{r} - (d_{r}^{\prime} + \Delta d)\right|\right)^2}{2 \sigma^2}\right),
                \end{equation}
                \end{small}
                \( d_{r} \) and \( d_{r}^{\prime} \) represent the depth attributes of the reference point $\textbf{x}$ respectively analogous to \( z \) and \( d \) in \cref{label:equation-projection}, they denote the distance of the sampled point from the ego-vehicle and the distance of the ray's corresponding object from the ego-vehicle, respectively. It should be noted that \( d_{r} \) is derived from the predicted bin depth distribution \( D_{t} \) and is transformed into relative depth, while \( \Delta d \) represents the bin interval. The parameter \( \sigma \) serves as an adjustment factor for \( d_{r} \) and \( d_{r}^{\prime} \), enabling fine-tuning of the depth matching tolerance. By default, \( \sigma \) is set to 2, offering a balance between strict and lenient matching. This design takes into account both the attributes of bin depth distribution and the state of the sampled point, enabling a more accurate modeling of 3D spatial information. 
                
            \subsubsection{Sparse Temporal Modeling}
                \begin{figure}[tb]
    \centering
    \includegraphics[width=0.45\textwidth]{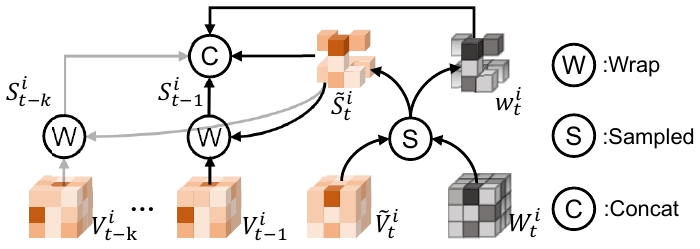}
    \vspace*{-0.1cm}
    \caption{ 
    \textbf{Illustration of Sparse Temporal Fusion.} We implement temporal fusion using a parallel strategy in a sparse manner, focusing only on modeling the sampled features.
    }
    \label{figure:temporal-fusion}
    \vspace{-1em}
\end{figure}

                \label{label:Long-Term Modeling}
                
                To integrate long-term historical information into the feature representation, at each stage \( i \), we maintain a streaming history memory bank that adheres to the first-in, first-out rule to dynamically fuse information using a parallel strategy. Inspired by the SlowFast~\cite{slowfast} and considering the redundancy in 3D space, we decouple the non-empty and empty regions into long-term and short-term streams, respectively. The long-term stream models the non-empty region to capture long-term dynamic information, while the short-term stream focuses on the empty region, modeling the overall 3D space in a short-term manner.
                \begin{table*}[tb]
    \setlength{\tabcolsep}{0.015\linewidth}
    \centering
    \scriptsize
  \scalebox{1.0}{
  \begin{tabular}{l|ccc|ccc|>{\columncolor{Gray}}ccc}
    \toprule
    Method & Backbone & Input Size & Epochs & RayIoU\textsubscript{1m}($\%$)$\uparrow$  & RayIoU\textsubscript{2m}($\%$)$\uparrow$  & RayIoU\textsubscript{4m}($\%$)$\uparrow$  & RayIoU($\%$)$\uparrow$  & Memory(G)$\downarrow$  \\
    \midrule
    BEVFormer$^{\ast}$~\cite{bevformer}       & R101   & 1600 $\times$ 900 & 24  & 26.1 & 32.9 & 38.0  & 32.4 & 24.1   \\
    RenderOcc$^{\ast}$~\cite{renderocc}       & Swin-B & 1408 $\times$ 512 & 12  & 13.4 & 19.6 & 25.5  & 19.5 & 17.5   \\
    BEVDet-Occ$^{\ast}$~\cite{bevdet}         & R50    & 704 $\times$ 256  & 90  & 23.6 & 30.0 & 35.1  & 29.6 & 10.2   \\
    FB-Occ$^{\ast}$~\cite{fbbev}              & R50    & 704 $\times$ 256  & 90  & 26.7 & 34.1 & 39.7  & 33.5 & 9.6    \\
    SparseOcc (16f)~\cite{sparseocc}                & R50    & 704 $\times$ 256  & 24  & 29.1 & 35.8 & 40.3  & 35.1 & 23.7    \\
    COTR$^{\dag}$~\cite{cotr}                          & R50    & 704 $\times$ 256  & 24  & 36.3 & 41.7 & 45.1  & 41.0 & 18.3     \\
    OPUS-L~\cite{opus}                    & R50    & 704 $\times$ 256  & 100 & 34.7 & 42.1 & 46.7  & 41.2 & 12.1      \\
    \midrule
    STCOcc (ours)                   & R50  & 704 $\times$ 256  & 36  & 36.2 & 42.7 & 46.4  & 41.7 & \textbf{7.7} \\
    STCOcc (ours)                    & R50  & 1408 $\times$ 512  & 36  & \textbf{36.9} & \textbf{42.8} & \textbf{46.7}  & \textbf{42.1} & 8.9 \\
    \bottomrule
  \end{tabular}
  }
  \caption{\textbf{Comparison of RayIoU ($\%$) performance on the Occ3D-nus dataset.} ${\ast}$ indicates models trained with camera mask, $^{\dag}$ denotes that official code was utilized to retrain the model.}
  \label{table:main_results_rayiou_on_occ3d}
\end{table*}
                \begin{table*}[t]
  \small
  \setlength{\tabcolsep}{0.0030\linewidth}
  \centering
  \begin{tabular}{l | cc | c | >{\columncolor{Gray}}c | c c c c c c c c c c c c c c c c c c}
      \toprule
      Method
      & \rotatebox{90}{Backbone}
      & \rotatebox{90}{Input Size}
      & \rotatebox{90}{Memory(G)}
      & \rotatebox{90}{mIoU($\%$)}
      & \rotatebox{90}{\textcolor{otherscolor}{$\blacksquare$} others} 
      & \rotatebox{90}{\textcolor{barriercolor}{$\blacksquare$} barrier} %
      & \rotatebox{90}{\textcolor{bicyclecolor}{$\blacksquare$} bicycle} %
      & \rotatebox{90}{\textcolor{buscolor}{$\blacksquare$} bus} %
      & \rotatebox{90}{\textcolor{carcolor}{$\blacksquare$} car} %
      & \rotatebox{90}{\textcolor{constructcolor}{$\blacksquare$} cons. veh.} %
      & \rotatebox{90}{\textcolor{motorcolor}{$\blacksquare$} motor.} %
      & \rotatebox{90}{\textcolor{pedestriancolor}{$\blacksquare$} pedes.} %
      & \rotatebox{90}{\textcolor{trafficcolor}{$\blacksquare$} tfc. cone} %
      & \rotatebox{90}{\textcolor{trailercolor}{$\blacksquare$} trailer} %
      & \rotatebox{90}{\textcolor{truckcolor}{$\blacksquare$} truck} %
      & \rotatebox{90}{\textcolor{driveablecolor}{$\blacksquare$} drv. surf.} %
      & \rotatebox{90}{\textcolor{otherflatcolor}{$\blacksquare$} other flat} %
      & \rotatebox{90}{\textcolor{sidewalkcolor}{$\blacksquare$} sidewalk} %
      & \rotatebox{90}{\textcolor{terraincolor}{$\blacksquare$} terrain} %
      & \rotatebox{90}{\textcolor{manmadecolor}{$\blacksquare$} manmade} %
      & \rotatebox{90}{\textcolor{vegetationcolor}{$\blacksquare$} vegetation} \\ %
      \midrule
      
      BEVFormer~\cite{bevformer}            & R101  & 1600$\times$900   &24.1 & 26.9 & 5.9 & 37.8 & 17.9 & 40.4 & 42.4 & 7.4 & 23.9 & 21.8 & 21.0 & 22.4 & 30.7 & {55.4} & 28.4 & 36.0 & 28.1 & 20.0 & 17.7 \\
      CTF-Occ~\cite{occ3d}                  & R101  & 1600$\times$900   & - & 28.5 & 8.1 & 39.3 & 20.6 & 38.3 & 42.2 & 16.9 & 24.5 & 22.7 & 21.1 & 23.0 & 31.1 & 53.3 & 33.8 & 38.0 & 33.2 & 20.8 & 18.0 \\
      TPVFormer~\cite{tpvformer}            & R101  & 1600$\times$900   & 28.9& 27.8 & 7.2 & 38.9 & 13.7 & 40.8 & 45.9 & 17.2 & 20.0 & 18.9 & 14.3 & 26.7 & 34.2 & 55.7 & 35.5 & 37.6 & 30.7 & 19.4 & 16.8 \\
      OSP$^{\ast}$~\cite{osp}               & R101  & 1600$\times$900   & 20.7 & 41.2 & 10.9 & 49.0 & 27.7 & 50.2 & 55.9 & 22.9 & 31.0 & 30.9 & 30.3 & 35.6 & 31.2 & 82.1 & 42.6 & 51.9 & 55.1 & 44.8 & 38.2 \\
      SparseOcc (8f)~\cite{sparseocc}       & R50   & 704$\times$256    & 14.3& 30.9 & 10.6 & 39.2 & 20.2 & 32.9 & 43.3 & 19.4 & 23.8 & 23.4 & 29.3 & 21.4 & 29.3 & 67.7 & 36.3 & 44.6 & 40.9 & 22.0 & 21.9 \\
      FB-Occ$^{\ast}$~\cite{fb-occ}         & R50   & 704$\times$256    & 9.6 & 39.1 & 13.6 & 44.7 & 27.0 & 45.4 & 49.1 & 25.2 & 26.3 & 27.9 & 27.8 & 32.3 & 36.8 & 80.1 & 42.8 & 51.2 & 55.1 & 42.2 & 37.5 \\
      ViewFormer$^{\ast}$~\cite{viewformer} & R50   & 704$\times$256    & -& 41.9 & 12.9 & 50.1 & 27.9 & 44.6 & 52.9 & 22.4 & 29.6 & 28.0 & 29.2 & 35.2 & 39.4 & 84.7 & 49.4 & 57.4 & 59.7 & 47.4 & 40.6 \\
      COTR$^{\ast}$~\cite{cotr}             & R50   & 704$\times$256    & 18.3 & 44.5 & 13.3 & 52.1 & 31.9 & 46.0 & 55.6 & \textbf{32.6} & 32.8 & 30.4 & \textbf{34.1} & 37.7 & 41.8 & \textbf{84.5} & 46.2 & \textbf{57.6} & \textbf{60.7} & \textbf{51.9} & \textbf{46.3} \\
      \hline
      STCOcc$^{\ast}$ (ours) & R50 & 704$\times$256  & \textbf{7.7} & 44.6 & 15.3 & \textbf{52.9} & 31.6 & 46.4 & 55.9 & 31.5 & 32.6 & 32.1 & 34.5 & 39.5 & 42.5 & 83.6 & 47.8 & 56.4 & 60.1 & 50.8 & 44.8  \\
      STCOcc$^{\ast}$ (ours) & R50 & 1408$\times$512  & 8.9 & \textbf{45.0} & \textbf{15.2} & 52.3 & \textbf{32.2} & \textbf{50.5}& \textbf{56.5} & 31.7 & \textbf{33.9} & \textbf{33.4} & 33.8 & \textbf{38.9} & \textbf{44.9} & 83.9 & \textbf{47.4} & 57.1 &60.1 & 50.6 & 42.7\\ 
      
  \bottomrule
  \end{tabular}
  \caption{\textbf{Comparison of mIoU ($\%$) performance on the Occ3d-nus dataset.} ${\ast}$ indicates models trained with camera mask.}
  \label{table:main_results_miou_on_occ3d}
\end{table*}
                
                For clarity, we consider one stream as an example, as shown in \cref{figure:temporal-fusion}. Given the current refined voxel feature \( \tilde{V}_{t}^{i} \) and the occupied weights \( W^{i}_{t} \), we first apply a top-k sampling method to extract the seed feature \( \tilde{S}_{t}^{i} \in \mathbb{R}^{C \times N_{S}} \) and corresponding occupied state $w^{i}_{t} \in \mathbb{R}^{C_{s} \times N_{S}}$, where $C_{s}$ represents the embedding dimensions for occupied weights. Utilizing the position of \( S_{t}\) and the ego-pose transformation matrix \( T_{t}^{t-j} \) from frame \( t \) to frame \( t-j \), we can retrieve the corresponding historical feature \( S_{t-j}^{i} \). We then concatenate all the corresponding historically sampled features with $w_{t}^{i}$, and subsequently apply MLP to fuse the information along the channel dimension:
                \begin{equation}
                    S^{i}_{t} = MLP( [ \tilde{S}_{t}^{i}, S_{t-1}^{i}, \cdots, S_{t-k}^{i}, w_{t}^{i}] ).
                \end{equation}
                
                Finally, we add \( S^{i}_{t} \) to the corresponding position in \( \tilde{V}_{t}^{i} \). By default, at each stage, we use \( T_{i} \) frames to model the long-term stream and \(\frac{T_{i}}{2}\) frames to model the short-term stream. At each stage \( i \), the fusion output $V_{t}^{i}$ is appended to the memory queue, and a BEV representation $B_{t}^{i}$ is stored for the subsequent frame of OA-TSA. This strategy provides a long-term perspective for recurrent temporal modeling and mitigates the gradient vanishing issue that is commonly encountered in previous recurrent temporal modeling methods \cite{bevformer}.

    \subsection{Loss}
        We compute the occupancy loss for each stage, adopting the Scene-Class Affinity Loss (\(\mathcal{L}_{scal}\)) from MonoScene \cite{monoscene}. This loss is applied to both semantic and geometric predictions to ensure accurate scene understanding. Given the sparsity and class imbalance inherent in 3D scenes, we also utilize a weighted focal loss~\cite{fb-occ} combined with the Lovasz loss~\cite{lovasz}. The loss for each stage $i$ is formulated as follows:

        \begin{equation}
            \mathcal{L}_{occ}^{i} = \mathcal{L}_{scal}^{geo}+\mathcal{L}_{scal}^{sem}+\mathcal{L}_{focal}+\mathcal{L}_{lov}.
        \end{equation}
     
        When considering the scene flow, we utilize the L1 loss $\mathcal{L}_{1}$ to supervise only the foreground voxel, The overall loss function is formulated as follow:
   
        \begin{equation}
            \mathcal{L} = \lambda_{f}\mathcal{L}_{1} + \mathcal{L}_{depth} + \sum_{i=1}^{N}w_{i}\times\mathcal{L}_{occ}^{i},
        \end{equation}        
        where $\mathcal{L}_{depth}$ is the cross-entropy loss used to supervise the depth network. $w_{i}$ is computed by $\frac{1}{2^{N-i}}$.
        
\section{Experiments}

    \subsection{Experimental Setup}
        \paragraph{Dataset.} To evaluate the performance of our model, we utilize the Occ3D-nus dataset \cite{occ3d} for assessing 3D occupancy prediction and the OpenOcc dataset \cite{occnet} for evaluating scene flow prediction. Both datasets are derived from the NuScenes dataset \cite{nuscenes}, encompassing 600 outdoor scenes for training, 150 for validation, and 150 for testing. Since the Occ3D-nus dataset does not provide scene flow labels, we have omitted the flow head when using this dataset. Additionally, since NuScenes does not provide depth labels, we follow previous methods~\cite{bevdepth, bevstereo} by projecting LIDAR points onto the image plane to serve as depth labels.

        \paragraph{Metric.} The mean Intersection-over-Union (mIoU) is a prevalent metric for assessing occupancy prediction performance in the Occ3D-nus dataset. However, this metric only accounts for the visible area at the current moment and may not fully reflect the model's completion capabilities~\cite{sparseocc}. Consequently, we also employ Ray-Based IoU (RayIoU)~\cite{sparseocc} to evaluate occupancy prediction performance. RayIoU is computed at three distance thresholds: 1, 2, and 4 meters. The final ranking metric is obtained by averaging the results across these thresholds. Additionally, we assess the performance of scene flow prediction using the mean absolute velocity error (mAVE)~\cite{occnet} across defined categories (e.g., car, truck). The mAVE is calculated for the set of true positives within a query ray threshold of 2 meters.

        \paragraph{Implementation Details.} We adopt the depth network from BEVStereo~\cite{bevstereo} and configure our model with three processing stages. Within each stage, the number of layers in the OA transformer is set to 2. The amount of historical information preserved for each stage is 16, 8, and 4, respectively. Unless specified otherwise, all models use the AdamW optimizer~\cite{adamw} with a global batch size of 16. Moreover, because our sampling strategy results in variable GPU memory costs, we report the maximum value in our results.
        
    \begin{table}[tp]
    \setlength{\tabcolsep}{0.007\linewidth}
    \centering
    \scriptsize
  \scalebox{0.95}{
  \begin{tabular}{l|ccc|c>{\columncolor{Gray}}cc}
    \toprule
    Method      & Sup. &Backbone  & Input Size & RayIoU($\%$)$\uparrow$   & mAVE$\downarrow$ & Mem(G)$\downarrow$\\
    \midrule
    OccNeRF-C~\cite{occnerf}        & C  & R101 &1600$\times$900  & 21.6  & 1.53 & - \\
    OccNeRF-L                       & L  & R101 &1600$\times$900  & 31.7  & 1.59 & -\\
    RenderOcc~\cite{renderocc}      & L  & R101 &1600$\times$900  & 36.7  & 1.63 & -\\
    Let Occ Flow~\cite{let-occ-flow}& C+L& R101 &1408$\times$512  & 40.5  & 1.45 & -\\
    OccNet~\cite{occnet}            & 3D & R101 &1600$\times$900  & 39.7  & 1.61 & -\\
    BEVFormer$^{\dag}$~\cite{bevformer}      & 3D & R50  &1600$\times$900  & 28.1   &1.12 & 26.0\\
    FB-Occ$^{\dag}$~\cite{fb-occ}            & 3D & R50  &704$\times$256   & 32.3   &0.83 & 11.1\\
    SparseOcc$^{\dag}$~\cite{sparseocc}      & 3D & R50  &704$\times$256   & 33.4   &0.87 & 15.8\\
    \midrule
    STCOcc (ours)                   & 3D & R50 &704$\times$256  & \textbf{40.8} &\textbf{0.44} & \textbf{8.7}   \\  
    \bottomrule
  \end{tabular}
  }
  \vspace{-0.5em}
  \caption{\textbf{Comparison of RayIoU ($\%$) and mAVE performance on the OpenOcc~\cite{occnet} dataset.} C and L denote Camera and Lidar supervision. $^{\dag}$ denotes that we utilize the offical code and add the flow head to produce the results.}
  \label{table:main_results:main_results_rayiou_mave_on_openocc}
  
\end{table}
    \subsection{Main Results}

        \paragraph{Main Results on Occ3D-nus.} As shown in \cref{table:main_results_rayiou_on_occ3d} and \cref{table:main_results_miou_on_occ3d}. We compare our method with previous state-of-the art methods on 3D occupancy task. Our methods achieves the state of the art performance 41.7$\%$ in RayIoU and 44.6$\%$ in mIoU, which is particularly noteworthy given the significantly lower training costs of 7.7GB, as opposed to the high training costs associated with COTR (18.3 GB) and OPUS-L (12.1 GB). Furthermore, since our spatial refinement process is dependent on the image size, we resize the input to 1408$\times$512 and achieve improved results in terms of RayIoU and mIoU. 
        \vspace{-0.5em}
        \paragraph{Main Results on OpenOcc.} As demonstrated in \cref{table:main_results:main_results_rayiou_mave_on_openocc}, we conducted experiments on the OpenOcc dataset to assess the performance of our model in terms of occupancy and scene flow. Our approach, which employs a smaller backbone (ResNet-50) and a reduced image input size ($704 \times 256$), achieves RayIoU scores of 40.8$\%$ and mAVE of 0.44. These results surpass those of both OccNet~\cite{occnet} (which uses ResNet-101 with an input size of $1600 \times 900$) and Let Occ Flow~\cite{let-occ-flow} (which also uses ResNet-101 with an input size of $1408 \times 512$). 
    
    \subsection{Ablation Study}   
        To investigate the impact of various modules, we perform ablation experiments on the OpenOcc dataset~\cite{occnet}. It is important to note that the ablation experiments were conducted on half of the training dataset and then evaluated on the full validation set. Specifically, we used the first 300 sequences to constitute half of the training dataset. 
        \vspace{-0.5em}
        \paragraph{The Effectiveness of Each Component.}
        \begin{table}[tp]
    \resizebox{\linewidth}{!}{

        \begin{tabular}{cc|cc|ccc}
        \toprule
        \multicolumn{2}{c|}{Spatial} & \multicolumn{2}{c|}{Temporal} & \multicolumn{3}{c}{Metric} \\
        \midrule
        
        OA-SCA & SROP  & STF & OA-TSA & RayIoU($\%$)$\uparrow$ & mAVE$\downarrow$ & Mem(G)$\downarrow$ \\
        \midrule
                  &            &             &           & 35.1   & 1.32 & \textbf{5.5}       \\
        \ding{51} &            &             &           & 35.7   & 1.27 & 5.7       \\
        \ding{51} &  \ding{51} &             &           & 36.0   & 1.20 & 5.7        \\
        \ding{51} &  \ding{51} & \ding{51}   &           & 38.0   & 0.69 & 8.6          \\
        \rowcolor[HTML]{EFEFEF}
        \ding{51} & \ding{51}  & \ding{51}   & \ding{51} & \textbf{38.4}   & \textbf{0.63} & 8.7        \\
    
        \bottomrule
        \end{tabular}

    }
    \caption{\textbf{Ablation study on the each component.} SROP refers to the Self-Recursive Occupancy Predictor, OA-SCA refers to the Occupancy-Aware Spatial Cross Attention, OA-TSA refers to the Occlusion-Aware Temporal Self-Attention, and STF refers to the Sparse Temporal Fusion.}
    \label{tab:ablation_compo}
\end{table}
        \begin{figure}[t!]
    \centering
    \begin{subfigure}{0.18\textwidth}
        \centering
        \includegraphics[width=0.90\linewidth]{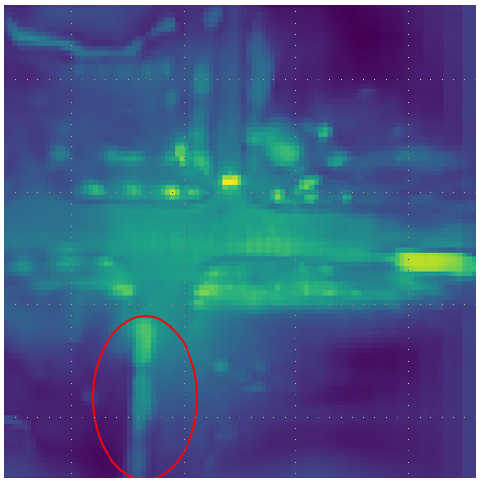}
        \caption{W OA-SCA.}
        \label{figure:w-OA-SCA}
    \end{subfigure}
    \centering
    \begin{subfigure}{0.18\textwidth}
        \centering
        \includegraphics[width=0.90\linewidth]{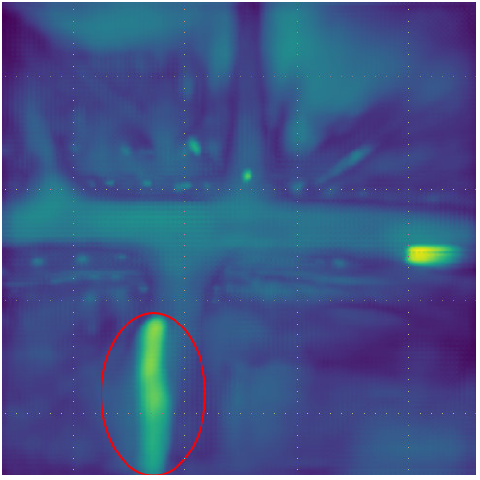}
        \caption{W/O OA-SCA.}
        \label{figure:wo-OA-SCA}
    \end{subfigure}
    \vspace{-0.5em}
    \caption{\textbf{Ablation on the OA-SCA module.} We visualize the features after refinement with and without the OA-SCA module.}
    \label{label:ablation-oa-sca}
\end{figure}

        \begin{figure*}[t]
    \centering
    \includegraphics[width=0.70\textwidth]{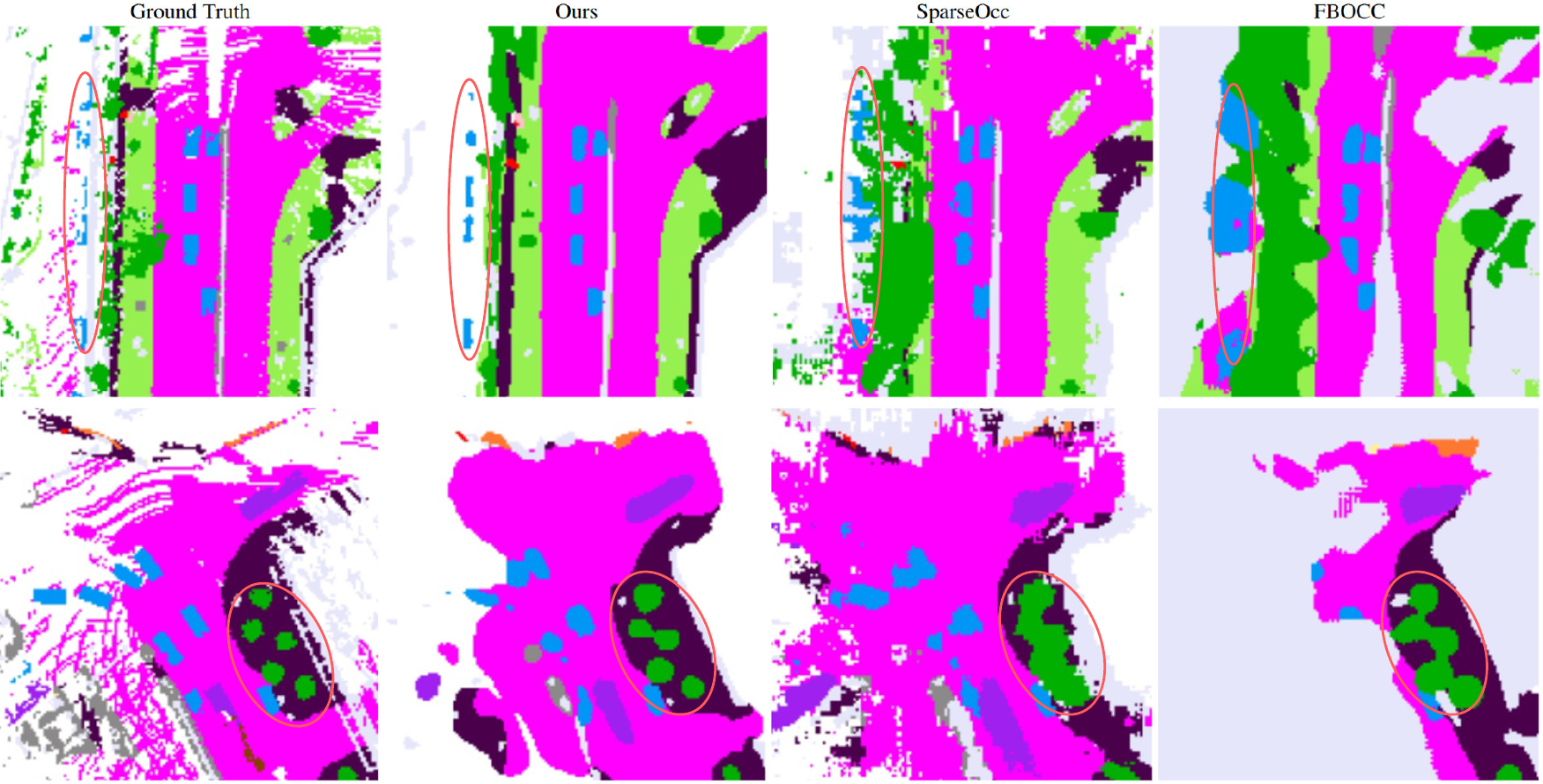}
    \vspace{-1em}
    \caption{ 
    \textbf{Qualitative results on Occ3d-nus validation set.} As depicted in the red circle, our method delivers detailed predictions for objects such as cars and trucks, while also offering clear boundary delineations for structures like buildings and vegetation.
    }
    \label{figure:visualization}
\end{figure*}
 
        In \cref{tab:ablation_compo}, we demonstrate the effectiveness of each component in our model. For the baseline, we omit the OA-SCA and make the occupancy predictor independent in each stage, similar to previous one-off selection methods~\cite{sparseocc, occnet}. This baseline achieves a RayIoU of 35.1$\%$ and a mAVE of 1.32 with a memory cost of 5.5GB. Integrating the OA-SCA into the baseline results in a 1.7$\%$ increase in RayIoU and a 3.7$\%$ increase in mAVE, with an additional memory cost of only 0.2GB. Introducing the SROP further enhances the model's performance without incurring any additional memory costs. Utilizing long-term temporal fusion and the OA-TSA, we achieve a 6.3$\%$ increase in RayIoU and a 45.5$\%$ increase in mAVE, with a memory cost of only 1.9GB. 
       
        \paragraph{The Effectiveness of Sparse Temporal Modeling.}
        In \cref{table:ablation-temporal-modeling}, we compare several representations in temporal modeling. The BEV modeling approach~\cite{solofusion}, while saving computation when modeling long-term history, sacrifices spatial information of the 3D space, resulting in poorer performance in occupancy prediction tasks. Our sparse modeling approach outperforms voxel-level modeling approach~\cite{fb-occ} in both effectiveness and computational cost. Furthermore, in \cref{tab:ablation-oa-tsa}, we compare our proposed Occupancy-Aware Temporal Self Attention (OA-TSA) with the vanilla Temporal Self Attention (TSA) \cite{bevformer}. Our method achieves superior performance in RayIoU and mAVE metrics, attributing this success to the guidance provided by occupied state modeling.
        
        \paragraph{The Effectiveness of Occupancy-Aware Spatial Cross Attention.} In \cref{label:ablation-oa-sca}, we compare the features refined by the OA-SCA module with those refined without it. It is evident that the OA-SCA module renovates the features, providing a more accurate geometric representation of the 3D scene, which is crucial for precise spatial modeling. Moreover, the OA-SCA module selectively enhances the foreground objects in the 3D scene, significantly improving the model's discriminability. Furthermore, in \cref{tab:ablation-oa-sca}, we compare the vanilla Spatial Cross Attention (SCA) \cite{bevformer} and the Depth-Aware Spatial Cross Attention (DA-SCA) proposed by FB-BEV \cite{fbbev} with our OA-SCA. It is observed that, due to inaccurate spatial modeling, neither SCA nor DA-SCA significantly improves performance. In contrast, Our explicit state-based modeling approach leverages the occupied state to accurately capture detailed spatial information.

    \begin{table}[tp]
    \setlength{\tabcolsep}{0.035\linewidth}
    \centering
    \scriptsize
  \scalebox{1.0}{
  \begin{tabular}{l|c|ccc}
    \toprule
    Representation & Frame & RayIoU($\%$)$\uparrow$ & mAVE$\downarrow$ & Mem(G)$\downarrow$ \\
    \midrule
    BEV~\cite{solofusion}       & 8 &   36.8 & 0.81 & \textbf{6.7} \\
    Voxel~\cite{fb-occ}     & 8 &   37.2 & 0.73 & 7.8  \\
    \rowcolor[HTML]{EFEFEF}
    Sparse (ours)    & 8 &   \textbf{37.5} & \textbf{0.71} & 7.3  \\
    \midrule
    BEV       & 16 &  37.4 & 0.77 & \textbf{7.6} \\
    Voxel     & 16 &  38.0 & 0.64 & 9.8 \\
    \rowcolor[HTML]{EFEFEF}
    Sparse (ours)   & 16 &  \textbf{38.4} & \textbf{0.60} & 8.7  \\
    \bottomrule
  \end{tabular}
  }
  \vspace{-0.5em}
  \caption{\textbf{Ablation on the Sparse Temporal Fusion.} We compare the traditional representations of BEV and voxel to our sparse modeling approach across various frame numbers.}
  \label{table:ablation-temporal-modeling}
  \vspace{-1em}
\end{table}
    \begin{table}[!t]
    \centering
	\begin{minipage}{0.26\textwidth}
		\centering
		\makeatletter\def\@captype{table}\makeatother
		\label{}
        \scalebox{0.75}{
		\begin{tabular}{lcc}
            \toprule
            Method    & RayIoU($\%$) & mAVE   \\
            \midrule
            W/O TSA~\cite{bevformer}    & 37.9  & 0.69   \\
            TSA        & 38.0  & 0.67       \\
            \midrule
            \rowcolor[HTML]{EFEFEF}
            OA-TSA     & \textbf{38.3} & \textbf{0.63} \\
            \bottomrule
		\end{tabular} 
        }
        \vspace{-0.5em}
        \caption{\textbf{Ablation on the OA-TSA Module.}}
        \label{tab:ablation-oa-tsa}
	\end{minipage}
    \hfill
        \begin{minipage}{0.20\textwidth}
    		\centering
    		\makeatletter\def\@captype{table}\makeatother
            \scalebox{0.65}{
    		\begin{tabular}{lc}
                \toprule
                Method    & RayIoU($\%$)   \\
                \midrule
                W/O SCA~\cite{bevformer} & 37.6    \\
                SCA                      & 37.5   \\
                DA-SCA~\cite{fbbev}      & 37.7   \\
                \midrule
                \rowcolor[HTML]{EFEFEF}
                OA-SCA                   & \textbf{38.3} \\
                \bottomrule
    		\end{tabular} 
            }
            \vspace{-0.5em}
            \caption{\textbf{Ablation on the OA-SCA Module.}} 
            \label{tab:ablation-oa-sca}
    	\end{minipage}
    
\vspace{-5pt}
\end{table}

    \subsection{Visualizations}

        In \cref{figure:visualization}, we present the BEV visualizations on the Occ3D-nus validation set. In comparison to implicit learning-based approaches~\cite{sparseocc, fb-occ}, our explicit state-based method produces clearer boundaries for objects such as cars, buildings, and vegetation.

\section{Conclusions}
    We propose an explicit state-based modeling approach to capture detailed geometric information in 3D space and integrate long-term dynamic information effectively. Our proposed STCOcc framework incorporates occlusion-aware mechanisms to enhance 3D features in both spatial and temporal aspects, thereby achieving better performance in 3D occupancy and flow prediction. The results demonstrate the efficacy of our paradigm, underscoring its strong potential for applications in downstream tasks.

\section*{Acknowledgement}
    This research was supported by the National Natural Science Foundation of China (No. U23B2060, No.62088102), and the Youth Innovation Team of Shaanxi Universities.

{
    \small
    \bibliographystyle{ieeenat_fullname}
    \bibliography{main}
}


\end{document}